\documentclass[logo]{dsme}

\usepackage[authoryear, sort&compress, round]{natbib}
\title{\mpx{}: Mixed Precision Training for JAX}

\usepackage{authblk}

\author[1]{Alexander Gräfe}
\author[1]{Sebastian Trimpe}

\renewcommand{\shortauthors}{Gräfe et al.}

\affil[1]{Institute for Data Science in Mechanical Engineering (DSME), RWTH Aachen University}

\codelabel{\faGithub} 
\codeurl{https://github.com/Data-Science-in-Mechanical-Engineering/mixed_precision_for_JAX}
\codeurlshort{mixed\_precision\_for\_JAX}

\usepackage{amsmath}
\usepackage{graphicx}
\usepackage{listings}
\usepackage{enumitem}
\usepackage{mdframed}

\usepackage{tikz}
\usepackage{pgfplots}
\pgfplotsset{compat=1.18}

\usepackage{caption}
\usepackage{subcaption}
\usepackage{siunitx}

\DeclareCaptionType[fileext=los,placement={!ht}]{example}

\newcommand{\mpx}{\textsc{MPX}}

\makeatletter
\def\PYG@reset{\let\PYG@it=\relax \let\PYG@bf=\relax%
    \let\PYG@ul=\relax \let\PYG@tc=\relax%
    \let\PYG@bc=\relax \let\PYG@ff=\relax}
\def\PYG@tok#1{\csname PYG@tok@#1\endcsname}
\def\PYG@toks#1+{\ifx\relax#1\empty\else%
    \PYG@tok{#1}\expandafter\PYG@toks\fi}
\def\PYG@do#1{\PYG@bc{\PYG@tc{\PYG@ul{%
    \PYG@it{\PYG@bf{\PYG@ff{#1}}}}}}}
\def\PYG#1#2{\PYG@reset\PYG@toks#1+\relax+\PYG@do{#2}}

\@namedef{PYG@tok@w}{\def\PYG@tc##1{\textcolor[rgb]{0.97,0.97,0.97}{##1}}}
\@namedef{PYG@tok@err}{\def\PYG@tc##1{\textcolor[rgb]{0.64,0.00,0.00}{##1}}\def\PYG@bc##1{{\setlength{\fboxsep}{\string -\fboxrule}\fcolorbox[rgb]{0.94,0.16,0.16}{1,1,1}{\strut ##1}}}}
\@namedef{PYG@tok@x}{\def\PYG@tc##1{\textcolor[rgb]{0.00,0.00,0.00}{##1}}}
\@namedef{PYG@tok@c}{\let\PYG@it=\textit\def\PYG@tc##1{\textcolor[rgb]{0.56,0.35,0.01}{##1}}}
\@namedef{PYG@tok@cm}{\let\PYG@it=\textit\def\PYG@tc##1{\textcolor[rgb]{0.56,0.35,0.01}{##1}}}
\@namedef{PYG@tok@cp}{\let\PYG@it=\textit\def\PYG@tc##1{\textcolor[rgb]{0.56,0.35,0.01}{##1}}}
\@namedef{PYG@tok@c1}{\let\PYG@it=\textit\def\PYG@tc##1{\textcolor[rgb]{0.56,0.35,0.01}{##1}}}
\@namedef{PYG@tok@cs}{\let\PYG@it=\textit\def\PYG@tc##1{\textcolor[rgb]{0.56,0.35,0.01}{##1}}}
\@namedef{PYG@tok@k}{\let\PYG@bf=\textbf\def\PYG@tc##1{\textcolor[rgb]{0.13,0.29,0.53}{##1}}}
\@namedef{PYG@tok@kc}{\let\PYG@bf=\textbf\def\PYG@tc##1{\textcolor[rgb]{0.13,0.29,0.53}{##1}}}
\@namedef{PYG@tok@kd}{\let\PYG@bf=\textbf\def\PYG@tc##1{\textcolor[rgb]{0.13,0.29,0.53}{##1}}}
\@namedef{PYG@tok@kn}{\let\PYG@bf=\textbf\def\PYG@tc##1{\textcolor[rgb]{0.13,0.29,0.53}{##1}}}
\@namedef{PYG@tok@kp}{\let\PYG@bf=\textbf\def\PYG@tc##1{\textcolor[rgb]{0.13,0.29,0.53}{##1}}}
\@namedef{PYG@tok@kr}{\let\PYG@bf=\textbf\def\PYG@tc##1{\textcolor[rgb]{0.13,0.29,0.53}{##1}}}
\@namedef{PYG@tok@kt}{\let\PYG@bf=\textbf\def\PYG@tc##1{\textcolor[rgb]{0.13,0.29,0.53}{##1}}}
\@namedef{PYG@tok@o}{\let\PYG@bf=\textbf\def\PYG@tc##1{\textcolor[rgb]{0.81,0.36,0.00}{##1}}}
\@namedef{PYG@tok@ow}{\let\PYG@bf=\textbf\def\PYG@tc##1{\textcolor[rgb]{0.13,0.29,0.53}{##1}}}
\@namedef{PYG@tok@p}{\let\PYG@bf=\textbf\def\PYG@tc##1{\textcolor[rgb]{0.00,0.00,0.00}{##1}}}
\@namedef{PYG@tok@n}{\def\PYG@tc##1{\textcolor[rgb]{0.00,0.00,0.00}{##1}}}
\@namedef{PYG@tok@na}{\def\PYG@tc##1{\textcolor[rgb]{0.77,0.63,0.00}{##1}}}
\@namedef{PYG@tok@nb}{\def\PYG@tc##1{\textcolor[rgb]{0.13,0.29,0.53}{##1}}}
\@namedef{PYG@tok@bp}{\def\PYG@tc##1{\textcolor[rgb]{0.20,0.40,0.64}{##1}}}
\@namedef{PYG@tok@nc}{\def\PYG@tc##1{\textcolor[rgb]{0.00,0.00,0.00}{##1}}}
\@namedef{PYG@tok@no}{\def\PYG@tc##1{\textcolor[rgb]{0.00,0.00,0.00}{##1}}}
\@namedef{PYG@tok@nd}{\let\PYG@bf=\textbf\def\PYG@tc##1{\textcolor[rgb]{0.36,0.21,0.80}{##1}}}
\@namedef{PYG@tok@ni}{\def\PYG@tc##1{\textcolor[rgb]{0.81,0.36,0.00}{##1}}}
\@namedef{PYG@tok@ne}{\let\PYG@bf=\textbf\def\PYG@tc##1{\textcolor[rgb]{0.80,0.00,0.00}{##1}}}
\@namedef{PYG@tok@nf}{\def\PYG@tc##1{\textcolor[rgb]{0.00,0.00,0.00}{##1}}}
\@namedef{PYG@tok@py}{\def\PYG@tc##1{\textcolor[rgb]{0.00,0.00,0.00}{##1}}}
\@namedef{PYG@tok@nl}{\def\PYG@tc##1{\textcolor[rgb]{0.96,0.47,0.00}{##1}}}
\@namedef{PYG@tok@nn}{\def\PYG@tc##1{\textcolor[rgb]{0.00,0.00,0.00}{##1}}}
\@namedef{PYG@tok@nx}{\def\PYG@tc##1{\textcolor[rgb]{0.00,0.00,0.00}{##1}}}
\@namedef{PYG@tok@nt}{\let\PYG@bf=\textbf\def\PYG@tc##1{\textcolor[rgb]{0.13,0.29,0.53}{##1}}}
\@namedef{PYG@tok@nv}{\def\PYG@tc##1{\textcolor[rgb]{0.00,0.00,0.00}{##1}}}
\@namedef{PYG@tok@vc}{\def\PYG@tc##1{\textcolor[rgb]{0.00,0.00,0.00}{##1}}}
\@namedef{PYG@tok@vg}{\def\PYG@tc##1{\textcolor[rgb]{0.00,0.00,0.00}{##1}}}
\@namedef{PYG@tok@vi}{\def\PYG@tc##1{\textcolor[rgb]{0.00,0.00,0.00}{##1}}}
\@namedef{PYG@tok@l}{\def\PYG@tc##1{\textcolor[rgb]{0.00,0.00,0.00}{##1}}}
\@namedef{PYG@tok@m}{\let\PYG@bf=\textbf\def\PYG@tc##1{\textcolor[rgb]{0.00,0.00,0.81}{##1}}}
\@namedef{PYG@tok@mf}{\let\PYG@bf=\textbf\def\PYG@tc##1{\textcolor[rgb]{0.00,0.00,0.81}{##1}}}
\@namedef{PYG@tok@mh}{\let\PYG@bf=\textbf\def\PYG@tc##1{\textcolor[rgb]{0.00,0.00,0.81}{##1}}}
\@namedef{PYG@tok@mi}{\let\PYG@bf=\textbf\def\PYG@tc##1{\textcolor[rgb]{0.00,0.00,0.81}{##1}}}
\@namedef{PYG@tok@il}{\let\PYG@bf=\textbf\def\PYG@tc##1{\textcolor[rgb]{0.00,0.00,0.81}{##1}}}
\@namedef{PYG@tok@mo}{\let\PYG@bf=\textbf\def\PYG@tc##1{\textcolor[rgb]{0.00,0.00,0.81}{##1}}}
\@namedef{PYG@tok@ld}{\def\PYG@tc##1{\textcolor[rgb]{0.00,0.00,0.00}{##1}}}
\@namedef{PYG@tok@s}{\def\PYG@tc##1{\textcolor[rgb]{0.31,0.60,0.02}{##1}}}
\@namedef{PYG@tok@sb}{\def\PYG@tc##1{\textcolor[rgb]{0.31,0.60,0.02}{##1}}}
\@namedef{PYG@tok@sc}{\def\PYG@tc##1{\textcolor[rgb]{0.31,0.60,0.02}{##1}}}
\@namedef{PYG@tok@sd}{\let\PYG@it=\textit\def\PYG@tc##1{\textcolor[rgb]{0.56,0.35,0.01}{##1}}}
\@namedef{PYG@tok@s2}{\def\PYG@tc##1{\textcolor[rgb]{0.31,0.60,0.02}{##1}}}
\@namedef{PYG@tok@se}{\def\PYG@tc##1{\textcolor[rgb]{0.31,0.60,0.02}{##1}}}
\@namedef{PYG@tok@sh}{\def\PYG@tc##1{\textcolor[rgb]{0.31,0.60,0.02}{##1}}}
\@namedef{PYG@tok@si}{\def\PYG@tc##1{\textcolor[rgb]{0.31,0.60,0.02}{##1}}}
\@namedef{PYG@tok@sx}{\def\PYG@tc##1{\textcolor[rgb]{0.31,0.60,0.02}{##1}}}
\@namedef{PYG@tok@sr}{\def\PYG@tc##1{\textcolor[rgb]{0.31,0.60,0.02}{##1}}}
\@namedef{PYG@tok@s1}{\def\PYG@tc##1{\textcolor[rgb]{0.31,0.60,0.02}{##1}}}
\@namedef{PYG@tok@ss}{\def\PYG@tc##1{\textcolor[rgb]{0.31,0.60,0.02}{##1}}}
\@namedef{PYG@tok@g}{\def\PYG@tc##1{\textcolor[rgb]{0.00,0.00,0.00}{##1}}}
\@namedef{PYG@tok@gd}{\def\PYG@tc##1{\textcolor[rgb]{0.64,0.00,0.00}{##1}}}
\@namedef{PYG@tok@ge}{\let\PYG@it=\textit\def\PYG@tc##1{\textcolor[rgb]{0.00,0.00,0.00}{##1}}}
\@namedef{PYG@tok@gr}{\def\PYG@tc##1{\textcolor[rgb]{0.94,0.16,0.16}{##1}}}
\@namedef{PYG@tok@gh}{\let\PYG@bf=\textbf\def\PYG@tc##1{\textcolor[rgb]{0.00,0.00,0.50}{##1}}}
\@namedef{PYG@tok@gi}{\def\PYG@tc##1{\textcolor[rgb]{0.00,0.63,0.00}{##1}}}
\@namedef{PYG@tok@go}{\let\PYG@it=\textit\def\PYG@tc##1{\textcolor[rgb]{0.00,0.00,0.00}{##1}}}
\@namedef{PYG@tok@gp}{\def\PYG@tc##1{\textcolor[rgb]{0.56,0.35,0.01}{##1}}}
\@namedef{PYG@tok@gs}{\let\PYG@bf=\textbf\def\PYG@tc##1{\textcolor[rgb]{0.00,0.00,0.00}{##1}}}
\@namedef{PYG@tok@ges}{\let\PYG@bf=\textbf\let\PYG@it=\textit\def\PYG@tc##1{\textcolor[rgb]{0.00,0.00,0.00}{##1}}}
\@namedef{PYG@tok@gu}{\let\PYG@bf=\textbf\def\PYG@tc##1{\textcolor[rgb]{0.50,0.00,0.50}{##1}}}
\@namedef{PYG@tok@gt}{\let\PYG@bf=\textbf\def\PYG@tc##1{\textcolor[rgb]{0.64,0.00,0.00}{##1}}}
\@namedef{PYG@tok@fm}{\def\PYG@tc##1{\textcolor[rgb]{0.00,0.00,0.00}{##1}}}
\@namedef{PYG@tok@vm}{\def\PYG@tc##1{\textcolor[rgb]{0.00,0.00,0.00}{##1}}}
\@namedef{PYG@tok@sa}{\def\PYG@tc##1{\textcolor[rgb]{0.31,0.60,0.02}{##1}}}
\@namedef{PYG@tok@dl}{\def\PYG@tc##1{\textcolor[rgb]{0.31,0.60,0.02}{##1}}}
\@namedef{PYG@tok@mb}{\let\PYG@bf=\textbf\def\PYG@tc##1{\textcolor[rgb]{0.00,0.00,0.81}{##1}}}
\@namedef{PYG@tok@pm}{\let\PYG@bf=\textbf\def\PYG@tc##1{\textcolor[rgb]{0.00,0.00,0.00}{##1}}}
\@namedef{PYG@tok@ch}{\let\PYG@it=\textit\def\PYG@tc##1{\textcolor[rgb]{0.56,0.35,0.01}{##1}}}
\@namedef{PYG@tok@cpf}{\let\PYG@it=\textit\def\PYG@tc##1{\textcolor[rgb]{0.56,0.35,0.01}{##1}}}

\def\PYGZus{\char`\_}

\makeatother

\begin{document}
\sloppy
\maketitle
\begin{abstract}
Mixed-precision training has emerged as an indispensable tool for enhancing the efficiency of neural network training in recent years. 
    Concurrently, JAX has grown in popularity as a versatile machine learning toolbox.
    However, it currently lacks robust support for mixed-precision training.
    We propose MPX, a mixed-precision training toolbox for JAX that simplifies and accelerates the training of large-scale neural networks while preserving model accuracy. 
    MPX seamlessly integrates with popular toolboxes such as Equinox and Flax, allowing users to convert full-precision pipelines to mixed-precision versions with minimal modifications. 
    By casting both inputs and outputs to half precision, and introducing a dynamic loss-scaling mechanism, MPX alleviates issues like gradient underflow and overflow that commonly arise in half precision computations. 
    Its design inherits critical features from JAX's type-promotion behavior, ensuring that operations take place in the correct precision and allowing for selective enforcement of full precision where needed (e.g., sums, means, or softmax). 
    MPX further provides wrappers for automatic creation and management of mixed-precision gradients and optimizers, enabling straightforward integration into existing JAX training pipelines.
    \mpx{}'s source code, documentation, and usage examples are available at \url{github.com/Data-Science-in-Mechanical-Engineering/mixed_precision_for_JAX}.
\end{abstract}

\section{Introduction}

In recent years, JAX~\citep{deepmind2020jax} has emerged as a popular tool for training neural networks. 
A standout feature of JAX is its just-in-time compilation, which significantly accelerates code execution on hardware accelerators such as GPUs and TPUs. 
Additionally, JAX supports scalable multi-device and multi-node training across computing clusters through user-friendly and efficient sharding mechanisms.

To extend JAX's capabilities, several libraries have been developed, including Equinox, Flax, and Optax~\citep{kidger2021equinox,flax2020github,deepmind2020jax}. 
Despite these advancements, one critical area where JAX falls short is in supporting mixed-precision training, a method essential for efficiently handling large models~\citep{mixed_precision_paper}. 
Although JMP~\citep{jmp} was previously available for this purpose, it is no longer maintained and cannot process arbitrary JAX PyTrees commonly used in modern neural-network toolboxes.

Mixed-precision training plays a pivotal role in optimizing the performance of large model training by executing both forward and backward passes in half-precision floating-point format (\qty{16}{\bit}). 
This technique significantly reduces memory requirements for intermediate activations and speeds up training by minimizing memory transfer overheads while utilizing specialized half-precision tensor cores (when supported by hardware), all without compromising accuracy~\citep{mixed_precision_paper}.

To facilitate the straightforward implementation of mixed-precision training in JAX, we introduce \mpx{}, a mixed-precision training toolbox tailored for JAX. 
\mpx{} builds upon the JMP implementation and leverages Equinox's flexibility to overcome limitations associated with handling arbitrary PyTrees. 
It further introduces gradient transformations and optimizer wrappers that enable seamless conversion from a full-precision pipeline to mixed precision with minimal modifications. The source code, documentation, and usage examples for \mpx{} are accessible at \url{github.com/Data-Science-in-Mechanical-Engineering/mixed_precision_for_JAX}.

Before explaining the core functionality of \mpx{}, we provide an overview of mixed-precision training principles. 
Subsequently, we demonstrate how \mpx{} can be integrated into existing workflows through sample code illustrating its main features.
Finally, we evaluate the performance of \mpx{} in the context of training vision transformers.

\section{Basics of mixed precision Training}
In this section, we summarize the foundational principles of mixed precision training as outlined in the original method~\citep{mixed_precision_paper}. 
Mixed precision training involves executing most computations during the forward and backward passes of a neural network using 16-bit floating-point numbers. 
This approach offers two primary advantages: 
\begin{itemize} 
    \item Reduced GPU Memory Usage: Mixed precision can decrease memory usage by approximately \SI{50}{\percent} compared to full precision. 
    \item Potential Speedup: This technique can lead to faster execution due to reduced memory access times and the utilization of specialized half-precision tensor cores available in TPUs or GPUs. 
\end{itemize}

\subsection{Loss Scaling}
One downside of using half-precision floating-point numbers is their decreased resolution. 
Consequently, small gradients may be rounded to zero, which can negatively impact training performance.

To mitigate these effects, loss scaling has been proposed. 
The main idea involves multiplying the loss by a factor greater than one and calculating the gradient of this scaled loss. 
Consequently, the gradients are also scaled by this factor, making them larger and allowing them to be represented more accurately by half-precision numbers.

However, selecting a constant scaling factor poses challenges. 
A small scaling factor might still result in gradients being rounded to zero, while an excessively large scaling factor can cause overflows where gradients become infinite. 
Additionally, different stages of training may benefit from varying scaling factors depending on the training state.

Dynamic loss scaling addresses this issue by adapting the scaling factor at runtime.
Its objective is to maximize the scaling factor for best accuracy with half-precision numbers while avoiding overflow. 
This is achieved through a simple heuristic that periodically increases the scaling factor and decreases it when gradients turn infinite.

In summary, mixed precision training with dynamic loss scaling involves the following steps:
\begin{enumerate}
    \item Calculate the prediction of the model in half-precision.
    \item Calculate the loss and scale it by a factor greater than 1.
    \item Calculate gradients of the scaled loss with respect to the model parameters.
    \item Converting gradients to float32.
    \item Divide them by the scaling factor to retrieve original gradients.
    \item Update the scaling factor:
    \begin{enumerate}[label=(\alph*)]
        \item If scaled gradients exceed float16's representable range (resulting in infinity), reduce the scaling and skip updating model parameters. 
        \item If scaled gradients remain within float16 range for an extended period, increase scaling.
    \end{enumerate}
    \item Update the model parameters.
\end{enumerate}

The primary goal of \mpx{} is to offer functionalities that streamline the implementation of all these steps in mixed precision training. 
In the following section, we provide an overview of these features.

\section{Implementation Details}

The \mpx{} library offers essential transformations for mixed precision training while preserving JAX's low-level flexibility. 
Similar to JMP~\citep{jmp}, our implementation's design leverages JAX's type-promotion behavior\footnote{\url{https://docs.jax.dev/en/latest/jep/9407-type-promotion.html}}, ensuring that operations are executed in the precision to which inputs and outputs have been cast—provided all constants in the function are positioned on the left side of the type-promotion lattice. 
Consequently, \mpx{} primarily focuses on casting function inputs and outputs to the desired precision.

Specifically, \mpx{} first defines functions for casting PyTrees (Section~\ref{sec:castingpytrees}), then utilizes these functions to cast both general functions (Section~\ref{sec:castingfunctions}) and gradient calculations (Section~\ref{sec:castinggradients}) into mixed precision. 
The gradient calculations include dynamic loss scaling (Section~\ref{sec:scaling}) to stabilize mixed precision training.
Moreover, \mpx{} provides wrappers around optimizers to exclude update steps where gradients are infinite (Section~\ref{sec:optimizer}).

\subsection{Transformations to Cast PyTrees}
\label{sec:castingpytrees}

\mpx{} provides several functions for casting arbitrary PyTrees to a desired floating-point data type. These functions include:
\texttt{cast\_tree(tree, dtype)}, \texttt{cast\_to\_half\_precision(x)}, \texttt{cast\_to\_float16(x)}, \texttt{cast\_to\_bfloat16(x)}, and \texttt{cast\_to\_float32(x)}. 
Each function inspects the leaves of the input PyTree, casting any leaf that is a JAX array with a floating-point type to the desired data type. 
Other leaves, such as integer arrays, remain unchanged. 
Excluding integer arrays is crucial to prevent accidental casting of random generator keys.

\subsection{Transformations to Cast Functions}
\label{sec:castingfunctions}

For casting entire functions, \mpx{} provides the transformation \texttt{cast\_function(func, dtype, return\_dtype=None)}. 
This transformation returns a new function that first applies the specified input data type to all inputs (via \texttt{cast\_tree(tree, dtype)}), then calls the original function, and finally casts the outputs to the specified (optional) return data type. 
Furthermore, \mpx{} includes \texttt{force\_full\_precision(func, return\_dtype)}, which ensures computations are carried out in full precision. 
This feature is particularly important for operations that are prone to overflow in \texttt{float16}.

\subsection{Automatic Loss Scaling}
\label{sec:scaling}

To facilitate the calculation of gradients via loss scaling, \mpx{} offers the class \texttt{DynamicLossScaling}, which extends \texttt{jmp.DynamicLossScaling}. 
This class manages dynamic loss scaling through the following methods:

\begin{itemize}
    \item \texttt{scale(x)}: Multiplies all floating-point leaves of the PyTree \texttt{x} by the current scaling factor.
    \item \texttt{unscale(x)}: Divides all floating-point leaves of \texttt{x} by the scaling factor and casts them to full precision.
    \item \texttt{adjust(grads\_finite)}: Updates the scaling factor based on the stability of the gradients, following the techniques described by~\cite{mixed_precision_paper}.
\end{itemize}

Because \texttt{DynamicLossScaling} inherits from \texttt{eqx.Module}, it is itself a PyTree. 
Consequently, it can be used within \texttt{jit}-compiled functions and also be used for multi-device and multi-node training using replicated sharding.


\subsection{Transformations to Cast Gradient Calculation}
\label{sec:castinggradients}

\mpx{} provides Equinox-compatible versions of gradient-calculation functions: \texttt{filter\_grad(func, scaling, has\_aux=False, use\_mixed\_precision=True)} and \texttt{filter\_value\_and\_grad(func, scaling, has\_aux=False, use\_mixed\_precision=True)}. 
These functions compute gradients using mixed precision and loss scaling (as described in Section\ref{sec:scaling}) while automatically adjusting the loss scaling value. 
They act as drop-in replacements for the corresponding Equinox functions, enabling users to reuse existing training pipelines with minimal modifications (cf. Section\ref{sec:eqxvsmpxmodelupdate}).

To use these functions, the model's forward pass and loss computation should be combined into a single function, which is then passed to \texttt{mpx.filter\_value\_and\_grad} in place of \texttt{eqx.filter\_value\_and\_grad}. 
The transformed function then operates as follows (Figure~\ref{fig:overview}):

\begin{enumerate}
    \item Cast all input arguments to half precision (\texttt{float16} or \texttt{bfloat16}).
    \item Execute the original function (model forward pass and loss).
    \item Scale the function's output by the current dynamic scaling factor.
    \item Apply \texttt{equinox.filter\_grad} to obtain scaled gradients.
    \item Unscale the gradients by dividing them by the scaling factor.
    \item Check whether the gradients are finite.
    \item Update the loss scaling based on the finiteness of gradients.
    \item Return a tuple containing:
    \begin{itemize}
        \item The updated \texttt{scaling} object.
        \item A boolean indicating if the gradients are finite (for the optimizer step).
        \item The computed gradients, whose format is a PyTree with the same format as the model parameters.
        \item Auxiliary values (if \texttt{has\_aux=True}).
    \end{itemize}
\end{enumerate}

\subsection{Optimizer}
\label{sec:optimizer}

Since \mpx{} produces gradients as PyTrees, they are directly compatible with Optax optimizers~\citep{deepmind2020jax}.
However, as previously discussed, it may be necessary to skip optimizer updates if the gradients become infinite due to loss scaling. 
The function \texttt{optimizer\_update(model, optimizer, optimizer\_state, grads, grads\_finite)} implements this logic by updating the model only when gradients are finite.

Thus, instead of calling \texttt{optimizer.update} followed by \texttt{eqx.apply\_updates}, as in a typical Equinox training pipeline, one can simply use \texttt{mpx.optimizer\_update}.

\section{Example}
\label{sec:eqxvsmpxmodelupdate}
In this section, we provide an example to demonstrate the changes required in a training pipeline for mixed precision training via \mpx{}.

For the largest part, the implementation of the model must not be changed.
As \mpx{} works with arbitrary PyTrees, every Toolbox that defines their model/parameters as PyTrees, like Flax~\citep{flax2020github} or Equinox~\citep{kidger2021equinox} can be used in conjunction with \mpx{}. 
Only parts of the model that contain critical operations with a high risk of overflows, must be transformed to full precision via \texttt{mpx.force\_full\_precision}.
However, many JAX and also Equinox already implement many critical functions in a numerically stable way by casting to full precision if necessary, so that only few changes are necessary.

Model updates via \mpx{} closely resemble those performed with Equinox (Example~\ref{ex:update}). 
Instead of calling \texttt{eqx.filter\_grad}, \texttt{mpx.filter\_grad} is used. Similarly, \texttt{mpx.optimizer\_update} replaces invoking the Optax optimizer followed by a model update.
The remainder of the training step—such as loading batches, sharding the model and batch—remains unchanged. 
All these steps can be compiled using tools like \texttt{eqx.filter\_jit}.

\begin{example}[h]
    \centering
    \begin{subfigure}[t]{0.98\textwidth}
        \centering
        \begin{pbox}[]{}
           \begin{Verbatim}[commandchars=\\\{\},codes={\catcode`\$=3\catcode`\^=7\catcode`\_=8\relax}]
\PYG{n}{grads} \PYG{o}{=} \PYG{n}{eqx}\PYG{o}{.}\PYG{n}{filter\PYGZus{}grad}\PYG{p}{(}\PYG{n}{loss}\PYG{p}{)(}\PYG{n}{model}\PYG{p}{,} \PYG{n}{batch}\PYG{p}{)}
\PYG{n}{updates}\PYG{p}{,} \PYG{n}{optimizer\PYGZus{}state} \PYG{o}{=} \PYG{n}{optimizer}\PYG{o}{.}\PYG{n}{update}\PYG{p}{(}
    \PYG{n}{grads}\PYG{p}{,} \PYG{n}{optimizer\PYGZus{}state}\PYG{p}{,} \PYG{n}{eqx}\PYG{o}{.}\PYG{n}{filter}\PYG{p}{(}\PYG{n}{model}\PYG{p}{,} \PYG{n}{eqx}\PYG{o}{.}\PYG{n}{is\PYGZus{}array}\PYG{p}{))}
\PYG{n}{model} \PYG{o}{=} \PYG{n}{eqx}\PYG{o}{.}\PYG{n}{apply\PYGZus{}updates}\PYG{p}{(}\PYG{n}{model}\PYG{p}{,} \PYG{n}{updates}\PYG{p}{)}
\end{Verbatim}

    \end{pbox}
        \caption{Full Precision via Equinox}
    \end{subfigure}\hspace{0.1cm}
    \begin{subfigure}[t]{0.98\textwidth}
        \centering
        \begin{pbox}[]{}
            \begin{Verbatim}[commandchars=\\\{\},codes={\catcode`\$=3\catcode`\^=7\catcode`\_=8\relax}]
\PYG{n}{loss\PYGZus{}scaling}\PYG{p}{,} \PYG{n}{grads\PYGZus{}finite}\PYG{p}{,} \PYG{n}{grads} \PYG{o}{=} \PYG{n}{mpx}\PYG{o}{.}\PYG{n}{filter\PYGZus{}grad}\PYG{p}{(}\PYG{n}{loss}\PYG{p}{,} \PYG{n}{loss\PYGZus{}scaling}\PYG{p}{)(}
    \PYG{n}{model}\PYG{p}{,} \PYG{n}{batch}\PYG{p}{)}
\PYG{n}{model}\PYG{p}{,} \PYG{n}{optimizer\PYGZus{}state} \PYG{o}{=} \PYG{n}{mpx}\PYG{o}{.}\PYG{n}{optimizer\PYGZus{}update}\PYG{p}{(}
    \PYG{n}{model}\PYG{p}{,} \PYG{n}{optimizer}\PYG{p}{,} \PYG{n}{optimizer\PYGZus{}state}\PYG{p}{,} \PYG{n}{grads}\PYG{p}{,}\PYG{n}{grads\PYGZus{}finite}\PYG{p}{)}
\end{Verbatim}

            \end{pbox}
            \caption{Mixed Precision via \mpx{}}
    \end{subfigure}
    \caption{Implementation of model updates via Equinox and \mpx{}. \mpx{} is designed such that its model update routine closely resembles the one of Equinox.}
    \label{ex:update}
\end{example}

\section{Evaluation}
We evaluate \mpx{} on two distinct machines: a mid-level desktop PC equipped with an AMD Ryzen 9 5950X, \qty{82}{\giga\byte} RAM, and an Nvidia GeForce RTX4070 GPU featuring \qty{12}{\giga\byte} VRAM; and a node from a high-performance cluster (CLAIX-2023) comprising Intel Xeon 8468 Sapphire CPUs, \qty{512}{\giga\byte} RAM, and four NVIDIA H100 GPUs connected via NVLink each with \qty{96}{\giga\byte} VRAM. 
On paper, the desktop PC's GPU maintains equivalent computation speeds for both half and full precision, whereas the H100 GPUs deliver double the speed for half precision compared to full precision.

We train vision transformers (ViT)~\citep{dosovitskiy2020image,steiner2021train}. On the desktop PC, the transformer features a size of 256 with residual blocks containing one hidden layer of 800 neurons. T
The cluster's transformer mirrors ViT-Base dimensions, with a feature size of 768 and residual blocks having one hidden layer of 3072 neurons.

The desktop PC trains the ViT on CIFAR100 while the high-performance cluster handles ImageNet1k, dividing each batch equally across GPUs using a data-parallel approach. We train without gradient accumulation, which could further optimize runtime and memory consumption.

We vary the number of batches and measure the time required for each training step across the entire dataset, excluding data loading times. 
RAM usage is measured by disabling automatic preallocation and setting "XLA\_PYTHON\_CLIENT\_PREALLOCATE" to "platform," minimizing JAX's VRAM consumption\footnote{\url{https://docs.jax.dev/en/latest/gpu_memory_allocation.html}}. 
For training speed assessment, both settings are turned off.
Due to technical reasons, we only measure VRAM consumption on the desktop PC as an accurate VRAM measurement was not possible on the high-performance cluster. 

\begin{figure}[h]
    \centering
    \begin{subfigure}[t]{0.9\linewidth}
    \begin{tikzpicture}
        \begin{axis}[
            width=0.8\textwidth,
            height=5cm,
            xlabel={Batch Size},
            ylabel={VRAM (\qty{}{\giga\byte})},
            legend pos=south east,
            grid=major,
            grid style={dashed,gray!30},
            xmin=0,
            ymin=0
        ]
        \addplot[
            color=blue,
            mark=o,
            thick
        ] coordinates {
            (128, 2.290)
            (256, 4.724)
            (512, 8.860)
        };
        \addlegendentry{Full Precision}

        \addplot[
            color=red,
            mark=square*,
            thick,
            dashed
        ] coordinates {
            (128, 1.52)
            (256, 2.418)
            (512, 4.900)
            (768, 7.054)
            (1024, 9.222)
        };
        \addlegendentry{Mixed Precision}
        \end{axis}
    \end{tikzpicture}
    \end{subfigure}

    \caption{Comparison of GPU VRAM consumed for full precision and mixed precision as a function of the number of batches on the desktop PC.}
    \label{fig:memory}
\end{figure}
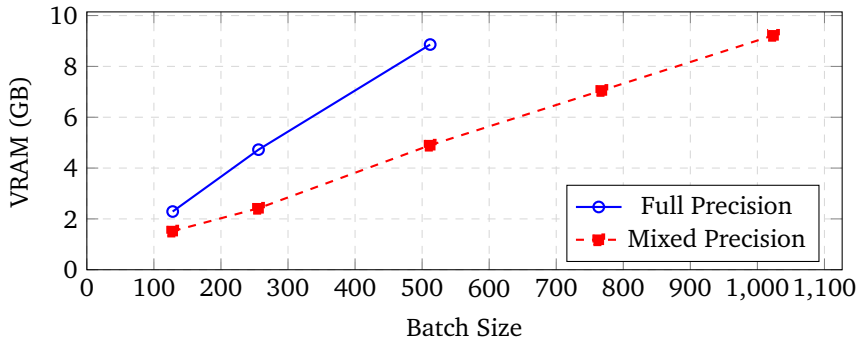

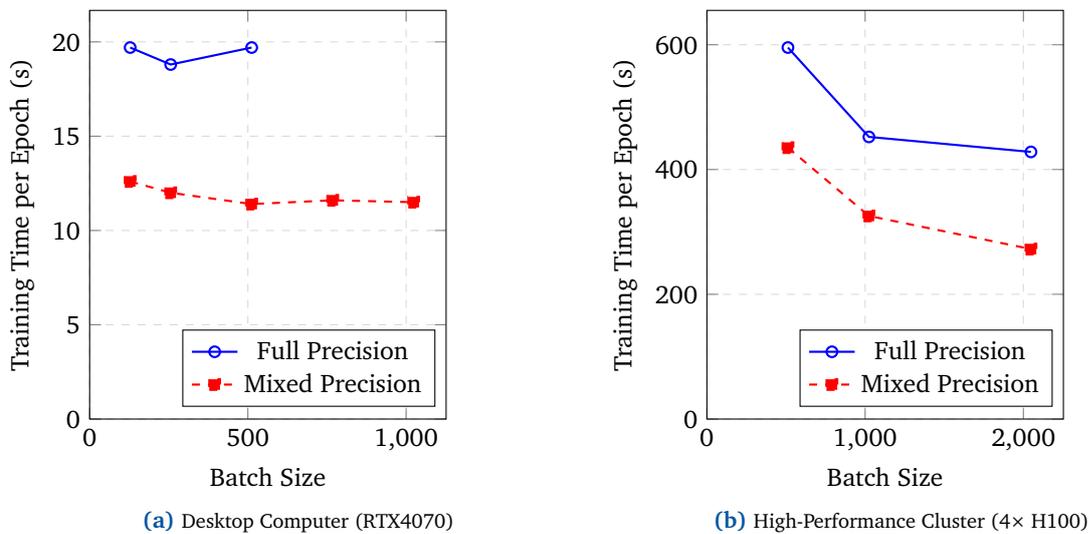
\begin{figure}[h]
    \centering
    \begin{subfigure}[t]{0.49\linewidth}
    \begin{tikzpicture}
        \begin{axis}[
            width=0.8\textwidth,
            height=7cm,
            xlabel={Batch Size},
            ylabel={Training Time per Epoch (s)},
            legend pos=south east,
            grid=major,
            grid style={dashed,gray!30},
            xmin=0,
            ymin=0,
        ]
        \addplot[
            color=blue,
            mark=o,
            thick
        ] coordinates {
            (128, 19.7)
            (256, 18.8)
            (512, 19.7)
        };
        \addlegendentry{Full Precision}

        \addplot[
            color=red,
            mark=square*,
            thick,
            dashed
        ] coordinates {
            (128, 12.6)
            (256, 12.0)
            (512, 11.4)
            (768, 11.6)
            (1024, 11.5)
        };
        \addlegendentry{Mixed Precision}
        \end{axis}
    \end{tikzpicture}
    \caption{Desktop Computer (RTX4070)}
    \end{subfigure}
    \begin{subfigure}[t]{0.49\linewidth}
    \begin{tikzpicture}
        \begin{axis}[
            width=0.8\textwidth,
            height=7cm,
            xlabel={Batch Size},
            ylabel={Training Time per Epoch (s)},
            legend pos=south east,
            grid=major,
            grid style={dashed,gray!30},
            xmin=0,
            ymin=0,
        ]
        \addplot[
            color=blue,
            mark=o,
            thick
        ] coordinates {
            (512, 595.5616071701049)
            (1024, 452.3)
            (2048, 428.24)
        };
        \addlegendentry{Full Precision}

        \addplot[
            color=red,
            mark=square*,
            thick,
            dashed
        ] coordinates {
            (512, 435.09)
            (1024, 325.61)
            (2048, 272.48)
        };
        \addlegendentry{Mixed Precision}
        \end{axis}
    \end{tikzpicture}
    \caption{High-Performance Cluster ($4\times$ H100)}
    \end{subfigure}
    \caption{Comparison of training step times for full precision and mixed precision as a function of the number of batches.}
    \label{fig:timing}
\end{figure}

Mixed precision training via \mpx{} reduces both RAM and processing times.
For the desktop PC, the RAM is reduced by factor $1.8\times$ and processing time by $1.7\times$.
The speedup is caused by reduced memory loading times as the RTX4070 streaming-multiprocessors feature no computing speedup for half precision.
The cluster features a similar behavior, where processing time is reduced by up to $1.57\times$.

These experiments demonstrate that mixed precision via \mpx{} allows to significantly speed up training.
Moreover, it significantly reduces VRAM usage.

\section{Conclusion}

\mpx{} offers an easy-to-use solution for mixed-precision training in JAX.
It is able to handle arbitrary PyTrees and provides a seamless integration with popular libraries like Equinox and Flax. 
At its core, \mpx{} allows casting functions to mixed precision and provides wrappers that allow to calculate gradients using mixed precision with dynamic loss scaling.
The evaluation demonstrates significant reductions in RAM usage and processing times. 
Overall, MPX presents a compelling toolbox for optimizing large-scale neural network training pipelines while maintaining flexibility and ease of use.

\section{Acknowledgements}
We thank Patrick Kidger for providing Equinox and Google DeepMind for JMP, which served as the foundation for this implementation.

We thank Lukas Wildberger for helpful discussions and suggestions.

The authors gratefully acknowledge the computing time provided to them at the NHR Center NHR4CES at RWTH Aachen University (project number p0021919). This is funded by the Federal Ministry of Education and Research, and the state governments participating on the basis of the resolutions of the GWK for national high performance computing at universities (\url{www.nhr-verein.de/unsere-partner}).

\bibliography{references}

\begin{thebibliography}{7}
\providecommand{\natexlab}[1]{#1}
\providecommand{\url}[1]{\texttt{#1}}
\expandafter\ifx\csname urlstyle\endcsname\relax
  \providecommand{\doi}[1]{doi: #1}\else
  \providecommand{\doi}{doi: \begingroup \urlstyle{rm}\Url}\fi

\bibitem[jmp()]{jmp}
Jmp: Jax mixed precision.
\newblock \url{https://github.com/google-deepmind/jmp}.
\newblock Accessed: 2024-06-09.

\bibitem[DeepMind et~al.(2020)DeepMind, Babuschkin, Baumli, Bell, Bhupatiraju,
  Bruce, Buchlovsky, Budden, Cai, Clark, Danihelka, Dedieu, Fantacci, Godwin,
  Jones, Hemsley, Hennigan, Hessel, Hou, Kapturowski, Keck, Kemaev, King,
  Kunesch, Martens, Merzic, Mikulik, Norman, Papamakarios, Quan, Ring, Ruiz,
  Sanchez, Sartran, Schneider, Sezener, Spencer, Srinivasan, Stanojevi\'{c},
  Stokowiec, Wang, Zhou, and Viola]{deepmind2020jax}
DeepMind, I.~Babuschkin, K.~Baumli, A.~Bell, S.~Bhupatiraju, J.~Bruce,
  P.~Buchlovsky, D.~Budden, T.~Cai, A.~Clark, I.~Danihelka, A.~Dedieu,
  C.~Fantacci, J.~Godwin, C.~Jones, R.~Hemsley, T.~Hennigan, M.~Hessel, S.~Hou,
  S.~Kapturowski, T.~Keck, I.~Kemaev, M.~King, M.~Kunesch, L.~Martens,
  H.~Merzic, V.~Mikulik, T.~Norman, G.~Papamakarios, J.~Quan, R.~Ring, F.~Ruiz,
  A.~Sanchez, L.~Sartran, R.~Schneider, E.~Sezener, S.~Spencer, S.~Srinivasan,
  M.~Stanojevi\'{c}, W.~Stokowiec, L.~Wang, G.~Zhou, and F.~Viola.
\newblock The {D}eep{M}ind {JAX} {E}cosystem, 2020.
\newblock URL \url{http://github.com/google-deepmind}.

\bibitem[Dosovitskiy et~al.(2020)Dosovitskiy, Beyer, Kolesnikov, Weissenborn,
  Zhai, Unterthiner, Dehghani, Minderer, Heigold, Gelly,
  et~al.]{dosovitskiy2020image}
A.~Dosovitskiy, L.~Beyer, A.~Kolesnikov, D.~Weissenborn, X.~Zhai,
  T.~Unterthiner, M.~Dehghani, M.~Minderer, G.~Heigold, S.~Gelly, et~al.
\newblock An image is worth 16x16 words: Transformers for image recognition at
  scale.
\newblock \emph{arXiv preprint arXiv:2010.11929}, 2020.

\bibitem[Heek et~al.(2024)Heek, Levskaya, Oliver, Ritter, Rondepierre, Steiner,
  and van {Z}ee]{flax2020github}
J.~Heek, A.~Levskaya, A.~Oliver, M.~Ritter, B.~Rondepierre, A.~Steiner, and
  M.~van {Z}ee.
\newblock {F}lax: A neural network library and ecosystem for {JAX}, 2024.
\newblock URL \url{http://github.com/google/flax}.

\bibitem[Kidger and Garcia(2021)]{kidger2021equinox}
P.~Kidger and C.~Garcia.
\newblock {E}quinox: neural networks in {JAX} via callable {P}y{T}rees and
  filtered transformations.
\newblock \emph{Differentiable Programming workshop at Neural Information
  Processing Systems 2021}, 2021.

\bibitem[Micikevicius et~al.(2017)Micikevicius, Narang, Alben, Diamos, Elsen,
  Garcia, Ginsburg, Houston, Kuchaiev, Venkatesh,
  et~al.]{mixed_precision_paper}
P.~Micikevicius, S.~Narang, J.~Alben, G.~Diamos, E.~Elsen, D.~Garcia,
  B.~Ginsburg, M.~Houston, O.~Kuchaiev, G.~Venkatesh, et~al.
\newblock Mixed precision training.
\newblock \emph{arXiv preprint arXiv:1710.03740}, 2017.

\bibitem[Steiner et~al.(2021)Steiner, Kolesnikov, Zhai, Wightman, Uszkoreit,
  and Beyer]{steiner2021train}
A.~Steiner, A.~Kolesnikov, X.~Zhai, R.~Wightman, J.~Uszkoreit, and L.~Beyer.
\newblock How to train your vit? data, augmentation, and regularization in
  vision transformers.
\newblock \emph{arXiv preprint arXiv:2106.10270}, 2021.

\end{thebibliography}

\end{document}